\pdfoutput=1

\documentclass[11pt]{article}

\usepackage[]{acl}

\usepackage{times}
\usepackage{latexsym}
\usepackage{url}
\usepackage{booktabs}
\usepackage{multirow}
\usepackage{multicol}
\usepackage{amsmath}
\usepackage{amssymb}
\usepackage{graphicx}
\usepackage{algorithm}
\usepackage[]{algpseudocode}

\DeclareMathOperator*{\argmin}{arg\,min}

\usepackage[T1]{fontenc}

\usepackage[utf8]{inputenc}

\usepackage{microtype}

\title{Revisiting the Practical Effectiveness of Constituency Parse Extraction from Pre-trained Language Models}

\author{Taeuk Kim \\
  Dept. of Computer Science \& Dept. of Artificial Intelligence \\
  Hanyang University, Seoul, South Korea \\
  \texttt{kimtaeuk@hanyang.ac.kr}}

\begin{document}
\maketitle
\begin{abstract}
Constituency Parse Extraction from Pre-trained Language Models (CPE-PLM) is a recent paradigm that attempts to induce constituency parse trees relying only on the internal knowledge of pre-trained language models.
While attractive in the perspective that similar to in-context learning, it does not require task-specific fine-tuning, the practical effectiveness of such an approach still remains unclear, except that it can function as a probe for investigating language models' inner workings.
In this work, we mathematically reformulate CPE-PLM and propose two advanced ensemble methods tailored for it, demonstrating that the new parsing paradigm can be competitive with common unsupervised parsers by introducing a set of heterogeneous PLMs combined using our techniques.
Furthermore, we explore some scenarios where the trees generated by CPE-PLM are practically useful.
Specifically, we show that CPE-PLM is more effective than typical supervised parsers in few-shot settings.
\end{abstract}

\section{Introduction}

With the increasing interest in the inner workings of pre-trained language models (PLMs; \citet{devlin-etal-2019-bert,liu2019roberta,radford2019language,conneau-etal-2020-unsupervised}),\footnote{We use the term \textit{pre-trained language models (PLMs)} to refer to the models that are based on Transformer \cite{vaswani2017attention} and pre-trained in a self-supervised manner, e.g., BERT \cite{devlin-etal-2019-bert} and its variants.} much work that attempts to explore the inherent knowledge embedded in the models has been recently proposed.
One of the main topics in this direction is to reveal whether PLMs understand syntactic knowledge of human language, usually represented as parse trees.
While a line of work (\citet{hewitt2019structural,chi-etal-2020-finding}) has investigated the existence of syntax in PLMs via structural probes with supervision from gold-standard parse trees, some studies \cite{Kim2020Are,kim-etal-2021-multilingual-chart-based,wu-etal-2020-perturbed} have found that one can extract reasonable parse structures directly from the patterns presented in PLMs' hidden representations or attention distributions \textit{even without extra fine-tuning}.
In other words, the studies have shown that PLMs implicitly store their understanding of syntactic knowledge in their parameters, and that such information can be easily reformulated into syntactic trees with almost no additional cost.

Although the aforementioned approach, dubbed \textbf{C}onstituency \textbf{P}arse \textbf{E}xtraction from \textbf{P}re-trained \textbf{L}anguage \textbf{M}odels (\textbf{CPE-PLM}; \citet{kim-etal-2021-multilingual-chart-based}), is undoubtedly a useful tool with many analytic uses \cite{rogers-etal-2020-primer}, it still remains a research question whether this algorithm can also work for practical purposes.
For instance, as CPE-PLM is free from fine-tuning of PLMs, it may be appealing in few-shot settings, akin to in-context learning for natural language understanding \cite{brown2020language}.
Moreover, there exists a potential that the new parsing paradigm can substitute the role of supervised or unsupervised parsers for the case where an NLP model requires a parse tree as input.

In this work, we focus on revealing the practical effectiveness of CPE-PLM.
Specifically, we first rewrite the procedure of CPE-PLM in a more rigorous form to clarify its working mechanism.
We then introduce two new ensemble algorithms tailored for it, i.e., \texttt{Greedy} and \texttt{Beam}, making its parsing performance competitive with that of unsupervised parsers \cite{kim-etal-2019-compound,zhu-etal-2020-the}.
We show that it is crucial to combine syntactic clues from heterogeneous PLMs for achieving comparable performance, and that this trend holds in not only English but also multilingual cases.

Equipped with the improved variants of CPE-PLM, as the next step, we investigate some scenarios in which their outputs (i.e., generated trees) can be practically utilized.
We show that (1) it is viable to introduce the trees from CPE-PLM as auxiliary data for improving Recurrent Neural Network Grammars (RNNG; \citet{dyer-etal-2016-recurrent,kim-etal-2019-unsupervised}), that (2) on classification with Tree LSTMs \cite{tai-etal-2015-improved}, the induced trees can replace gold-standard parses with a minimal loss, and that (3) CPE-PLM can be even more data-efficient than supervised parsers in few-shot settings.

\begin{figure*}[t!]
\begin{center}
\includegraphics[width=0.99\linewidth]{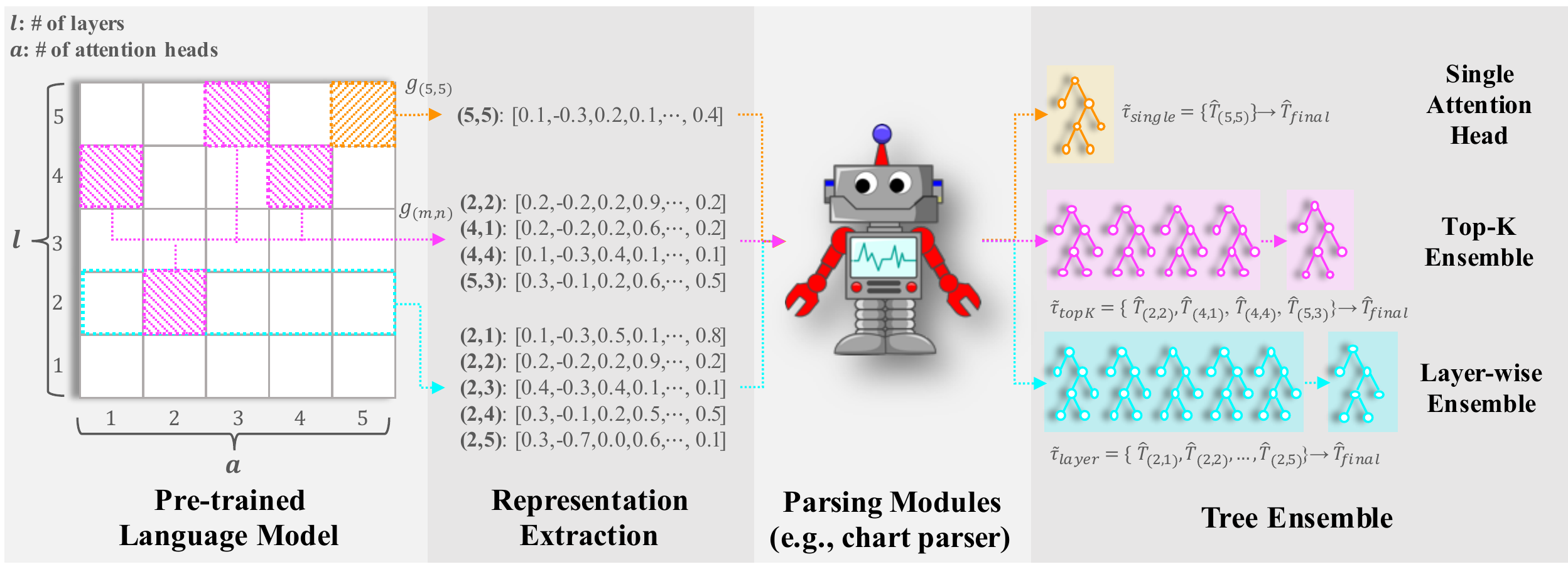}
\caption{Concept diagram explaining the procedure of CPE-PLM with various ensemble methods. Given a PLM that has $l=5$ Transformer layers, each of whose self-attention module consists of $a=5$ individual attention heads, an input sentence is inserted into the PLM to compute the model's attention maps. Then, we construct parse candidates using the information from each $g_{(m,m)}$ and the CKY algorithm. Here, the role of ensemble methods is to determine which trees to be engaged in the final prediction of the resulting parse tree.} 
\label{fig:figure1}
\end{center}
\end{figure*}

\section{Background and Related Work}

\subsection{Constituency Parse Extraction from Pre-trained Language Models} \label{subsec: constituency parse extraction from pre-trained language models}
 
The term ``\textbf{C}onstituency \textbf{P}arse \textbf{E}xtraction from \textbf{P}re-trained \textbf{L}anguage \textbf{M}odels (\textbf{CPE-PLM})'' coined by \citet{kim-etal-2021-multilingual-chart-based} represents a range of parsing methods \cite{marecek-rosa-2019-balustrades,rosa2019inducing,Kim2020Are,kim-etal-2021-multilingual-chart-based,li-etal-2020-heads} that aim to infer the parse tree of an input sentence by only exploiting the features obtained from PLMs.
In detail, the approaches belonging to this paradigm attempt to directly (i.e., without training) apply simple heuristics or existing parsing algorithms, such as top-down \cite{shen-etal-2018-straight} and chart-based \cite{kitaev-klein-2018-constituency} ones, on the hidden representations or attention maps retrieved from PLMs.
In the following, we illustrate the exact formulation of some representative methods.

As CPE-PLM does not demand more than \textit{frozen} PLMs as its ingredient, which means \textit{training-free}, it can be particularly useful when there are no resources available for training supervised parsers in terms of either (1) computing resources or (2) training data consisting of gold-standard annotations.
However, its use has been limited to analytic purposes in the literature, utilized as a tool for probing the inner workings of PLMs.
Our goal in this paper is therefore to investigate the utility of CPE-PLM in practical scenarios.
Among several options, we select the chart-based variant \cite{kim-etal-2021-multilingual-chart-based} as our baseline, which generally outperforms others.

\paragraph{Chart-based CPE-PLM.} 

\citet{kim-etal-2021-multilingual-chart-based} propose a method that combines PLMs with the chart parsing algorithm without extra training.
Formally, each tree candidate $T$ for an input sentence, $w_1, w_2, \dots, w_z$, is assigned a score $s_{tree}(T)$ that decomposes as $s_{tree}(T) = \sum_{(i,j) \in T} s_{span}(i, j)$, where $s_{span}(i, j)$ is a score for a constituent that is located between positions $i$ and $j$ in the sentence.
$s_{span}(i, j)$ is defined as follows:
\begin{equation*}
    \resizebox{.99\linewidth}{!}{
    $s_{span}(i, j) = 
    \begin{cases}
        s_{comp}(i,j) + \min_{i \le k < j}s_{split}(i, k, j) & \text{if} \ \ i < j \\
        0 & \text{if} \ \ i = j
    \end{cases}$},
\end{equation*}
where $s_{split}(i, k, j) = s_{span}(i, k) + s_{span}(k+1, j)$.
In other words, $s_{comp}(i,j)$ measures the compositionality of the span $(i,j)$ itself while $s_{split}(i, k, j)$ indicates how plausible it is to divide the span $(i,j)$ into two subspans $(i,k)$ and $(k+1, j)$. 
Note that every $s_{span}(i,j)$ can be easily calculated in a bottom-up fashion with the aid of the CKY algorithm \cite{cocke1969programming,kasami1966efficient,younger1967recognition}, once $s_{comp}(i,j)$ is properly defined.

Although the authors suggest two derivations for $s_{span}(i,j)$, in this work, the pair score function $s_{p}(\cdot,\cdot)$ is chosen as the main target of our interest, which is defined as follows:
\begin{equation*}
    \resizebox{.99\linewidth}{!}{
    $s_{p}(i,j) := \binom{j-i+1}{2}^{-1} \sum_{(w_x,w_y)\in pair(i,j)} f(g(w_x), g(w_y)),$}
\end{equation*}
where $pair(i,j)$ returns a set consisting of all combinations of two words from a span $(i,j)$, e.g., $pair(1,3) = \{(w_1,w_2), (w_1,w_3), (w_2,w_3)\}$, while $f(\cdot,\cdot)$ and $g(\cdot)$ are a distance measure function and representation extractor function respectively.
To realize $f(\cdot, \cdot)$ and $g(\cdot)$, the authors consider two sets of functions, $F$ and $G$.
Given $l$ as the number of layers in a PLM and $a$ as the number of attention heads per layer, $G$ refers to the set of functions $\{g_{(m,n)}|m=1,\dots,l, n=1,\dots,a\}$, each of which outputs the attention distribution of an input word computed by the $n^{th}$ attention head on the $m^{th}$ layer of the PLM.
$F$ is specified as $\{$\texttt{JSD}, \texttt{HEL}$\}$, where \texttt{JSD} and \texttt{HEL} correspond to the Jensen-Shannon and Hellinger distance.
Here, \texttt{HEL} is only considered for simplicity.

Intuitively, a series of the operations described so far can be understood as (1) splitting an attention map by rows (i.e., into attention distributions) for representing each word and (2) comparing similarities between the rows to gauge the syntactic proximity of the corresponding words.
Finally, chart-based CPE-PLM outputs $\hat{T}$, the tree that requires the lowest cost to build, as a prediction for the parse tree of the input sentence: $\hat{T} = \argmin_{T}s_{tree}(T)$.

\subsection{Ensemble Methods for CPE-PLM} \label{subsec: ensemble methods for CPE-PLM}

In Section \ref{subsec: constituency parse extraction from pre-trained language models}, we merely mentioned the case where only an element of $G$ is adopted for computing the scores used in CPE-PLM.
However, it is also feasible to employ more than just one, such that the method is provided with diverse information from different attention maps (i.e., $g_{(m,n)}$) to derive more reasonable parse trees. 
In fact, the previous work on CPE-PLM \cite{Kim2020Are,kim-etal-2021-multilingual-chart-based} already exploits such \textit{ensemble} strategies in an \textit{ambiguous} and \textit{implicit} manner to boost its performance.
As the introduction of those techniques can lead to a considerable gap in the final result, we claim that it is essential to clarify which ensemble method is employed and to develop more advanced ones.
Accordingly, we here \textit{explicitly} formulate the previous ensemble methods, which is one of our contributions in this work.

Figure \ref{fig:figure1} explains the process of applying ensemble methods to CPE-PLM.
Formally, let $\tau := \{\hat{T}_{(m,n)}|m=1,\dots,l, n=1,\dots,a\}$ denote a pool of all the possible tree predictions computed with CPE-PLM using every $g_{(m,n)}$. 
The objective of the ensemble methods is to derive the best parse prediction from a subset of $\tau$, denoted $\widetilde{\tau}$, so that it closely resembles the corresponding gold-standard tree.
Once $\widetilde{\tau}$ is decided, every $\hat{T}_{(m,n)}$ from $\widetilde{\tau}$ is converted into the form of \textit{syntactic distance} \cite{shen-etal-2018-straight} $\mathbf{d}_{(m,n)} \in \mathbb{R}^{z-1}$. \footnote{Note that $z$ is the number of words in the input sentence.}
Then, the resulting vectors are averaged to derive $\mathbf{d}_{final}$, which is finally restored to the tree form $\hat{T}_{final}$.\footnote{Refer to \citet{kim-etal-2021-multilingual-chart-based} for the exact procedure of converting a tree into a syntactic distance and vice versa.}
In the following, we illustrate the characteristics of each ensemble technique shown when determining $\widetilde{\tau}$.

\paragraph{Na\"ive Baseline: Single Attention Head.} 

The simplest way of implementing CPE-PLM is to utilize just a single attention head as a representative. 
To be specific, this baseline constructs $\widetilde{\tau}_{single} := \{\hat{T}_{(m^*,n^*)}| \forall m \forall n, val(g_{(m,n)}) \leq val(g_{(m^*,n^*)}) \}$, where $val(g_{(m,n)})$ indicates the performance of CPE-PLM on the \textit{validation} set, given $g_{(m,n)}$.
In other words, it directly outputs the parse $\hat{T}_{(m^*,n^*)}$ generated by the best function $g_{(m^*,n^*)}$.

\paragraph{Layer-wise Ensemble.} 

\citet{Kim2020Are} suggest to merge a group of trees that originated from the attention heads located in the same layer of a PLM.
Specifically, this ensemble method defines $\widetilde{\tau}_{layer}^{m} := \{\hat{T}_{(m,n)}|n=1,\dots,a\}$ to consider layer-specific information.
The best layer of the PLM (i.e., $m^*$) is also determined by its performance on the validation set.
The intuition behind this heuristic is that attention heads of the same layer might be complementary cooperative in grasping a linguistic concept, considering that particular layers of a PLM seem specialized to capture a specific aspect of linguistic knowledge \cite{tenney-etal-2019-bert,jawahar-etal-2019-bert,jo-myaeng-2020-roles}.

\paragraph{Top-K Ensemble.} 

\citet{kim-etal-2021-multilingual-chart-based} propose utilizing the top-K $g(\cdot)$ functions instead of only using the best, $g_{(m^*,n^*)}$.
First, a sorted set $\tau_{sorted} := \{\hat{T}_{(m^i,n^i)} | i,j \in \{1,\dots,l \times a\} \ \cap \ val(g_{(m^i,n^i)}) \geq val(g_{(m^j,n^j)}) \text{ whenever } i \leq j \}$ is specified as a variant of the set $\tau$.
That is, $\tau_{sorted}$ is identical to $\tau$ except that the elements of $\tau_{sorted}$ are arranged in descending order according to the validation performance of their corresponding functions $\{g_{(m^i,n^i)}\}$.
Then, the top-K ensemble method defines $\widetilde{\tau}_{topK}$ as the set consisting of the first $K$ elements of $\tau_{sorted}$.

\begin{figure}[t!]
\begin{center}
\includegraphics[width=0.99\linewidth]{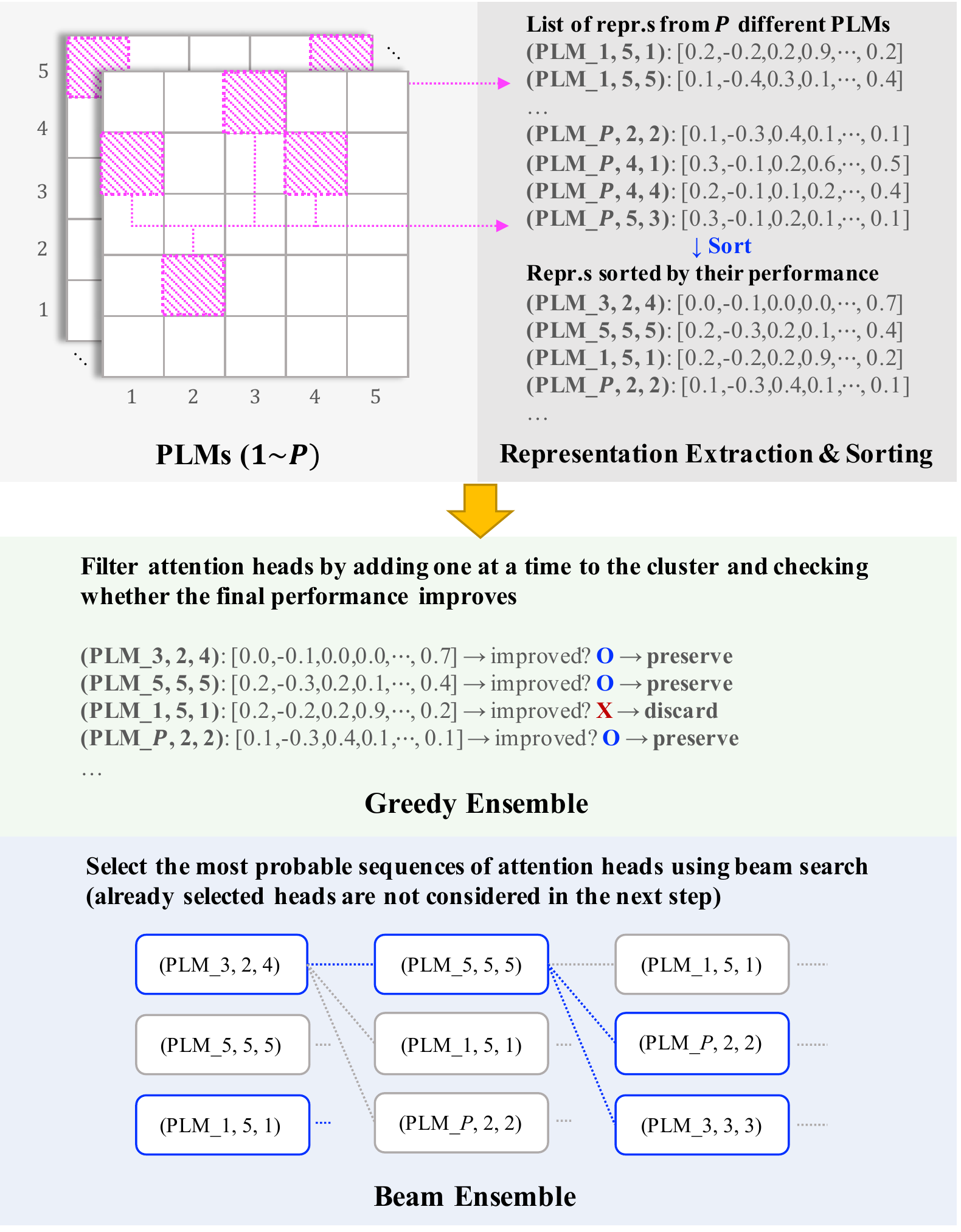}
\caption{Concept diagram explaining the operation of \texttt{Greedy} and \texttt{Beam} in multi-PLM environments.} 
\label{fig:figure2}
\end{center}
\end{figure}

\begin{algorithm}[ht]
\small
\caption{Greedy Ensemble Algorithm}
\label{alg: greedy ensemble} 
\begin{algorithmic}[1]
\State {$G_{sorted} := \{ g_{(m^i,n^i)} | i,j$$\in$$\{1,\dots,l$$\times$$a\}$$\cap$$val(g_{(m^i,n^i)})$ \hfill 
$\geq val(g_{(m^j,n^j)}) \text{ whenever } i \leq j \}$}
\State {$g_{(m^i,n^i)} := $ The $i^{th}$ element of $G_{sorted}$}
\State {$\hat{T}_{(m^i,n^i)} := $ A tree prediction generated using $g_{(m^i, n^i)}$}
\Function{Greedy}{$G_{sorted}$}
    \State {$G_{greedy}, \widetilde{\tau}_{greedy}, \mu \gets \{\}, \{\}, 0$}
    \For {$i=1,\dots,l \times a$}
        \State{$G_{greedy} \gets G_{greedy} \cup \{g_{(m^i,n^i)}\} $}
        \State{$\psi \gets val(G_{greedy})$}
        \If {$\psi > \mu$}
            \State {$\mu \gets \psi$}
            \State{$\widetilde{\tau}_{greedy} \gets \widetilde{\tau}_{greedy} \cup \{\hat{T}_{(m^i,n^i)}\} $} 
        \Else
            \State {$G_{greedy} \gets G_{greedy} \backslash \{g_{(m^i,n^i)}\}$}
        \EndIf
    \EndFor
    \State {\Return $\widetilde{\tau}_{greedy}$}
\EndFunction
\end{algorithmic}
\end{algorithm}

\section{Proposed Methods} \label{sec: proposed methods}

In this section, we additionally introduce two novel ensemble methods for CPE-PLM, which are more effective than the previous counterparts in improving parsing performance.
Furthermore, we propose to allow several PLMs to collaborate with each other to provide CPE-PLM with more diverse syntactic information (Figure \ref{fig:figure2}).

\paragraph{Greedy Ensemble.}

We first consider a method that collects every helpful attention head in a greedy fashion.
Unlike the previous ensemble methods which allow only a fixed number of attention heads to be engaged in the ensemble process, this sets no limit on the number of participants (i.e., attention heads), diversifying the source of syntactic clues.

Sticking to the notations defined in Section \ref{subsec: ensemble methods for CPE-PLM}, we specify the \texttt{Greedy} ensemble algorithm in Algorithm \ref{alg: greedy ensemble}.
Its core logic is to append one attention head at a time to the cluster and test whether each augmentation is beneficial for making progress in the final (validation) performance ($val(G_{greedy})$).

\paragraph{Beam Ensemble.}

Even though the greedy ensemble method is simple and effective, there still exists a need for exploring more diverse groups of attention heads that have the potential to show better performance than the group chosen by the greedy algorithm. 
To this end, inspired by the beam search algorithm widely adopted in the natural language generation (NLG) literature, we introduce the \texttt{Beam} ensemble method in Algorithm \ref{alg: beam ensemble}.

\begin{algorithm}[H]
\small
\caption{Beam Ensemble Algorithm}
\label{alg: beam ensemble} 
\begin{algorithmic}[1]
\State {$b := $ beam size}
\State {$G_{sorted} := \{ g_{(m^i,n^i)} | i,j$$\in$$\{1,\dots,l$$\times$$a\}$$\cap$$val(g_{(m^i,n^i)})$ \hfill 
$\geq val(g_{(m^j,n^j)}) \text{ whenever } i \leq j \}$}
\State {$g_{(m^i,n^i)} := $ The $i^{th}$ element of $G_{sorted}$}
\Function{Beam}{$G_{sorted}, b$}
    \State{$G_{beam}, e \gets \{\}, 0$}
    \For{$i=1,\dots,b$}
        \State{$G_{beam} \gets G_{beam} \cup \{\{g_{(m^i,n^i)}\}\}$}
    \EndFor
    \While{$e<b$}
        \State{$\hat{G}, \psi, e \gets \{\}, \{\}, 0$}
        \For{\textbf{each} $H \in G_{beam}$}
            \State{\resizebox{.79\linewidth}{!}{$\mu \gets$ The largest index $j$ given $\forall g_{(m^j,n^j)}$$\in$$H$.}}
            \For{$i=1,\dots,b$}
                \If{$\mu+i > l \times a$} 
                    \State{$e \gets e + 1$}
                    \State{\textbf{break}}
                \Else
                    \State{$H_i \gets H \cup \{g_{(m^{\mu+i},n^{\mu+i})}\} $}
                    \State{$\hat{G} \gets \hat{G} \cup \{H_i\}$}
                    \State{$\psi \gets \psi \cup \{val(H_i)\}$}
                \EndIf
            \EndFor
        \EndFor
        \State{\resizebox{.86\linewidth}{!}{$\psi_{topB} \gets $ The set consisting of the top $b$ elements of $\psi$.}}
        \State{$G_{beam} \gets \{H|H \in \hat{G} \cap val(H) \in \psi_{topB} \}$}
    \EndWhile
    \State{$G_{beam}^* \gets \{H^*|\forall H$$\in$$G_{beam}, val(H^*) \geq val(H)\}$}
    \State{$\widetilde{\tau}_{beam} \gets \{\hat{T}_{(m^i,n^i)})|\hat{T}_{(m^i,n^i)}$ is predicted using $g_{(m^i,n^i)} \in H^*$ ($H^*$ is the sole element of $G_{beam}^*$)$\}$}
    \State{\Return $\widetilde{\tau}_{beam}$}
\EndFunction
\end{algorithmic}
\end{algorithm}

Our beam ensemble algorithm is similar to one for NLG \cite{graves2012sequence,sutskever2014sequence} except that the beam search procedure is not stochastic, but determined by the order of the elements of $G_{sorted}$. Furthermore, the already selected attention heads are not considered in the next search step, unlike NLG which allows the same word to be generated twice or more (see Figure \ref{fig:figure2} for example).
The merit of \texttt{Beam} is that it can explore a wider range of potential paths that might not be covered by the greedy algorithm.

\paragraph{Extension to Multi-PLM Settings.}

Until now, we have assumed that CPE-PLM is only applicable for only a single PLM.
However, we propose for the first time extending its usage to the scenario in which multiple PLMs are available together.
In other words, we expand $\tau$ to $\tau_{multi} := \{\hat{T}_{(p,m,n)}|p\in\{1,\dots,P\}, m\in\{1,\dots,l\}, n\in\{1,\dots,a\}\}$, where $P$ is the number of the PLMs involved.
By doing so, it is expected that CPE-PLM can infer more accurate parse trees with the aid of diverse perspectives from heterogeneous PLMs.
We show in Section \ref{subsec: experiments on CPE-PLM with different ensemble methods} that this simple and intuitive extension leads to a significant improvement in the final performance.
It also has value in that it is one of the initial attempts in the literature to leverage different PLMs simultaneously.

\section{Experiments on Parsing}

In this chapter, we aim to validate the effectiveness of the proposed ensemble algorithms, i.e., \texttt{Greedy} and \texttt{Beam}, which enable CLE-PLM to have a higher potential of being properly adopted for downstream tasks.

\subsection{General Configurations}

\paragraph{Datasets.} 

To evaluate the parsing performance of CPE-PLM and other related models, we utilize the Penn Treebank dataset (PTB, \citet{marcus-etal-1993-building}) for English and the SPMRL \cite{seddah-etal-2013-overview} dataset for eight other languages, following \citet{kim-etal-2021-multilingual-chart-based}.\footnote {We use national codes to represent languages, i.e., en: English, eu: Basque, fr: French, de: German, he: Hebrew, hu: Hungarian, ko: Korean, pl: Polish, and sv: Swedish.}
We also adhere to the standard of the previous work for preprocessing the datasets.

\paragraph{Evaluation Metrics.} 

We utilize the \textit{unlabeled} sentence-level $F1$ score as a main metric to evaluate the extent to which induced trees resemble corresponding gold-standard trees.
It was originally introduced by \citet{shen2018neural,shen2019ordered}, becoming the de-facto standard in unsupervised parsing. 

\paragraph{Model Selection \& Hyperparameters.}

As mentioned in Section \ref{subsec: constituency parse extraction from pre-trained language models}, we build our approach upon chart-based CPE-PLM \cite{kim-etal-2021-multilingual-chart-based}. 
Moreover, we employ twelve English PLMs and four multilingual PLMs to provide syntactic information.\footnote{The list of PLMs we use is (1) English PLMs: BERT-base/large \cite{devlin-etal-2019-bert}, RoBERTa-base/large \cite{liu2019roberta}, ELECTRA-base/large \cite{Clark2020ELECTRA}, GPT2(-medium) \cite{radford2019language}, CTRL \cite{keskar2019ctrl}, BART \cite{lewis-etal-2020-bart}, XLNet \cite{yang2019xlnet}, (2) multilingual PLMs: MBERT, XLM \cite{NIPS2019_8928}, XLM-R(-large) \cite{conneau-etal-2020-unsupervised}.}
To handle various PLMs in an integrated manner, we use the \texttt{Transformers} library developed by HuggingFace \cite{wolf2019huggingface}.
We determine hyperparameters for the Top-K and \texttt{Beam} ensemble methods using grid search.  
In consequence, we use $K$=$20$ and $b$=$5$ for single PLM cases and $K$=$30$ and $b$=$30$ in multi-PLM settings.

\begin{table}[t!]
\scriptsize
    \centering
    \setlength{\tabcolsep}{0.2em}
    \begin{tabular}{l c c | c c c c c}
    \toprule
    \multirow{2}{*}{PLMs / Methods} & \multicolumn{2}{c}{Previous work} & \multicolumn{5}{|c}{Chart CPE-PLM w/ ensemble methods} \\
    & Top-down$^\dagger$ & Chart$^\ddagger$ & Single & Layer & Top-K & Greedy & Beam \\
    \midrule
   \bf Encoder-based & & & & & & & \\ 
    BERT-base & 32.4 & 42.7 & 34.1 & 35.3 & 42.5 & 43.0 & 42.8  \\
    BERT-large & 34.2 & 44.2 & 38.7 & 40.6 & 44.4 & 45.0 & 44.5 \\
    RoBERTa-base & 33.8 & 44.9 & 40.9 & 39.2 & 44.2 & 45.4 & 45.4 \\ 
    RoBERTa-large & 34.1 & 41.9 & 39.5 & 38.9 & 44.9 & 47.2 & 43.7  \\ 
    ELECTRA-base & - & - & 40.2 & 41.2 & 43.3 & 46.9 & 43.2 \\ 
    ELECTRA-large & - & - & \bf 44.3 & 41.3 & 46.6 & 47.9 & 47.2 \\ 
    \midrule
    \bf Decoder-based & & & & & & & \\ 
    GPT2 & 37.1 & 37.2 & 34.5 & 26.4 & 36.9 & 37.2 & 37.1 \\
    GPT2-medium & 39.4 & 38.4 & 38.0 & 28.2 & 38.2 & 38.0 & 40.8 \\
    CTRL & - & - & 35.7 & 28.7 & 44.4 & 45.8 & 44.9 \\
    \midrule 
    \bf Hybrid & & & & & & & \\
    BART-large & - & - & 37.5 & 32.6 & 39.8 & 37.5 & 38.5  \\
    XLNet-base & \textbf{40.1} & 46.4 & 36.7 & 39.7 & 46.0 & 47.0 & 46.7 \\
    XLNet-large & 38.1 & 46.4 & 39.5 & 38.9 & 45.7 & 47.2 & 46.8 \\
    \midrule
    \bf Multilingual & & & & & & & \\
    MBERT & - & 45.0 & 39.0 & 40.3 & 44.6 & 47.1 & 45.7 \\
    XLM & - & \textbf{47.7}& 41.9 & 42.1 & 47.1 & 47.5 & 47.1 \\
    XLM-R & - & 46.7 & 41.6 & \bf 44.2 & 46.5 & 48.5 & 47.4 \\
    XLM-R-large  & - & 44.6 & 40.7 & 36.7 & 44.3 & 46.8 & 46.9 \\
    \midrule
    \bf Multiple PLMs & & & & & & & \\ 
    Only multilingual & - & - & - & - & 49.6 & 51.9 & 49.8\\
    All models & - & - & - & - & \bf 50.4 & \bf 55.3 & \bf 55.7 \\
    \bottomrule
    \end{tabular}
    \caption{$F1$ scores of CPE-PLM on the PTB test set conditioned on the different combinations of PLMs and ensemble methods.
    We show that \texttt{Greedy} and \texttt{Beam} are more effective than baselines, and that we attain much more competitive scores in multiple PLM settings.
    The best score for each column is in \textbf{bold}.
    We also report numbers from two previous studies for reference. 
    $\dagger$: From \citet{Kim2020Are}. $\ddagger$: From \citet{kim-etal-2021-multilingual-chart-based}.}
    \label{table:table1}
\end{table}

\subsection{Verification of CPE-PLM's Performance} \label{subsec: experiments on CPE-PLM with different ensemble methods}

We first conduct experiments on the English PTB dataset using CPE-PLM, with the objective of comparing the effects of different PLMs and ensemble methods on the paradigm.
We also test the settings in which multiple PLMs are employed at the same time.
From Table \ref{table:table1}, we confirm that our \texttt{Greedy} and \texttt{Beam} algorithms are more effective than other techniques in most cases and that their impact is amplified when combined with multiple PLMs.
As a result, we succeed in achieving the state-of-the-art $F1$ score (55.7) on PTB in the CPE-PLM literature, improving by up to eight points compared against the previous best (47.7).
We also observe that Transformer encoder-based and multilingual PLMs are more attractive options for CPE-PLM.

\begin{table}[t!]
\scriptsize
\setlength{\tabcolsep}{0.4em}
\begin{center}
\begin{tabular}{l c c c c c c c}
\toprule
Models & $F_1$ & SBAR & NP & VP & PP & ADJP & ADVP \\
\midrule
\bf Unsupervised parsers \\
PRPN$^\dagger$ & 47.3 & 50 & 59 & 46 & 57 & 44 & 32 \\
ON-LSTM $^\dagger$ & 48.1 & 51 & 64 & 41 & 54 & 38 & 31 \\
Neural PCFG$^\dagger$ & 50.8 & 52 & 71 & 33 & 58 & 32 & 45 \\
Compound PCFG $^\dagger$ & 55.2 & \bf 56 & 74 & 41 & 68 & 40 & 52 \\
Neural L-PCFG$^\ddagger$ & 55.3 & 53 & 67 & \bf 48 & 65 & 49 & 58 \\
\midrule
\bf CPE-PLM (Ours) \\
XLM-R + Greedy & 48.5 & 46 & 69 & 29 & 62 & 48 & 73 \\
All PLMs + Greedy & 55.3 & 54 & \bf 75 & 36 & \bf 76 & \bf 50 & \bf 76 \\
All PLMs + Beam & \bf 55.7 & 53 & 74 & 42 & 75 & 46 & 72 \\
\bottomrule
\end{tabular}
\end{center}
\caption{Comparison of the best CPE-PLM variants with unsupervised parsers. 
We show that with the aid of \texttt{Greedy} and \texttt{Beam}, CPE-PLM becomes competitive with unsupervised PCFGs.
We also report recall scores on six phrasal categories in addition to $F1$ scores.
The best result for each column is in \textbf{bold}.
$\dagger$: From \citet{kim-etal-2019-compound}. $\ddagger$: From  \citet{zhu-etal-2020-the}.}
\label{table:table2}
\end{table}

Second, we take the best instances of CPE-PLM from Table \ref{table:table1} and compare them with a set of unsupervised parsers on PTB.
In particular, we consider PRPN \cite{shen2018neural}, ON \cite{shen2019ordered}, Neural PCFG, Compound PCFG \cite{kim-etal-2019-compound}, and Neural L-PCFG \cite{zhu-etal-2020-the} as baselines.
Note that all the models including CPE-PLM are evaluated on the same condition where we assume we have access to the validation set (for tuning hyperparameters), following the prior work \cite{kim-etal-2019-compound}.
From Table \ref{table:table2}, we show that with the introduction of \texttt{Greedy} and \texttt{Beam}, CPE-PLM becomes comparable to off-the-shelf unsupervised parsers in terms of $F1$.
Specifically, our CPE-PLM instance with \texttt{Beam} succeeds in achieving the best $F1$ score among all the candidates, and the variant with \texttt{Greedy} accomplishes the best recall scores on four phrasal categories (NP, PP, ADJP, and ADVP).
Based on these quantitative results, we claim that CPE-PLM is proper to be an alternative for unsupervised parsers in some cases.

\begin{table}[t!]
\tiny
\setlength{\tabcolsep}{0.4em}
\centering
\begin{tabular}{l c c c c c c c c c c }
\toprule
Models / Language & en & eu & fr & de & he & hu & ko & pl & sv & Avg. \\ 
\midrule
\bf Single PLM\\
MBERT \\
\hspace{1mm} Top-K ensemble & 44.6 & 39.3 & 35.9 & 35.9 & 37.8 & 33.2 & 47.5 & 51.1 & 32.6 & 39.8 \\
\hspace{1mm} Greedy ensemble & 47.1 & 40.2 & 36.9 & 37.5 & 38.6 & 30.2 & 49.1 & 52.4 & 31.9 & 40.4 \\
\hspace{1mm} Beam ensemble & 45.7 & 41.2 & 36.1 & 37.6 & 38.0 & 33.8 & 49.1 & 51.4 & 32.6 & 40.6 \\
XLM \\
\hspace{1mm} Top-K ensemble & 47.1 & 34.6 & 36.4 & 43.8 & 41.0 & 36.3 & 33.6 & 58.5 & 36.0 & 40.8  \\
\hspace{1mm} Greedy ensemble & 47.5 & 38.4 & 37.0 & 45.4 & 41.5 & 36.4 & 35.1 & 58.0 & 36.4 & 41.7  \\
\hspace{1mm} Beam ensemble & 47.1 & 38.7 & 36.8 & 43.6 & 41.9 & 36.3 & 35.2 & 56.8 & 36.2 & 41.4 \\
XLM-R \\
\hspace{1mm} Top-K ensemble & 46.5 & 39.5 & 35.8 & 37.5 & 40.1 & 36.6 & 49.8 & 52.7 & 32.8 & 41.3  \\
\hspace{1mm} Greedy ensemble & 48.5 & 39.4 & 36.1 & 39.0 & 40.3 & 36.5 & 50.8 & 53.5 & 33.1 & 41.9  \\
\hspace{1mm} Beam ensemble  & 47.4 & 39.0 & 35.3 & 37.9 & 39.7 & 37.0 & 50.2 & 53.9 & 32.7 & 41.5 \\
XLM-R-large \\
\hspace{1mm} Top-K ensemble & 44.3 & 37.2 & 29.7 & 36.3 & 35.8 & 31.0 & 45.5 & 44.7 & 27.6 & 36.9 \\
\hspace{1mm} Greedy ensemble & 46.8 & 39.5 & 32.9 & 40.1 & 37.0 & 34.0 & 46.4 & 47.1 & 31.0 & 39.4 \\
\hspace{1mm} Beam ensemble & 46.9 & 39.2 & 33.0 & 39.2 & 36.1 & 33.4 & 45.8 & 50.7 & 29.0 & 39.3 \\
\midrule
\bf Multiple PLMs \\
All models \\
\hspace{1mm} Top-K ensemble & 49.6 & 40.9 & 38.8 & 44.3 & 44.5 & 38.5 & 51.1 & 58.7 & 37.2 & 44.8 \\
\hspace{1mm} Greedy ensemble & \bf 51.9 & \bf 44.0 & \bf 41.9 & \bf 47.3 & \bf 48.1 & \bf 40.1 & \bf 53.7 & \bf 61.4 & \bf 39.0 & \bf 47.5 \\
\hspace{1mm} Beam ensemble & 49.8 & 42.7 & 40.4 & 47.0 & 45.9 & 39.4 & 53.4 & 60.8 & 38.2 & 46.4 \\
\bottomrule
\end{tabular}
\caption{CPE-PLM with three ensemble methods for nine languages.
We observe that it is optimal for every language to leverage \texttt{Greedy} on top of the integration of all the four multilingual PLMs considered. 
The best result for each column is in \textbf{bold}.}
\label{table:table3}
\end{table}

Finally, we extend the language domain of our experiments from English to eight other languages.
We exploit four multilingual PLMs that are capable of processing all the languages we consider, and each ensemble method is optimized for respective languages.
In Table \ref{table:table3}, we demonstrate that our ensemble methods are universally more effective than the top-K algorithm across different languages.
Furthermore, we confirm that CPE-PLM with \texttt{Greedy} attains the best performance in every case when operated on the combination of all the four PLMs and, showing 47.5 $F1$ score on average.

\section{Experiments on Downstream Tasks} \label{subsec: experiments on downstream tasks}

In the previous section, we demonstrated that CPE-PLM's performance can be significantly improved by introducing the techniques proposed in this work.
We now turn our attention towards its outputs (i.e., generated parse trees) and investigate the utility of such trees in two application scenarios where tree structures are taken as input. 

\subsection{Training (U)RNNG with Induced Trees}

Recurrent Neural Network Grammar (RNNG) \cite{dyer-etal-2016-recurrent} and its unsupervised variant (URNNG, \citet{kim-etal-2019-unsupervised}) are neural architectures which perform language modeling and parsing together.
In \citet{kim-etal-2019-compound}, the authors showed that training (U)RNNG with the trees generated by other unsupervised parsers results in a parsing model that is even better than the parsers which provided the trees used in training.
Following the previous work, we here examine whether the output trees from CPE-PLM can also function as meaningful signals for training (U)RNNG.
For our experiments, we acquire the best two instances from Table \ref{table:table1} (which accomplished 55.3 and 55.7 parsing $F1$ scores respectively) and use them as our pseudo parsers. 
We employ Compound PCFG \cite{kim-etal-2019-compound} as an unsupervised parser baseline.

\begin{table}[t!]
\scriptsize
\centering
\setlength{\tabcolsep}{0.6em}
\begin{tabular}{l r c}
\toprule
\textbf{From} \citet{kim-etal-2019-compound} (The \textit{best} over trials) & PPL ($\downarrow$) & $F_1$ ($\uparrow$) \\
\midrule
LSTM LM & 86.2 & $-$  \\
PRPN & 87.1 & 47.9 \\
 \hspace{3mm} Induced RNNG   & 95.3 & 47.8  \\
 \hspace{3mm} Induced URNNG  & 90.1 & 51.6   \\
ON & 87.2 & 50.0\\
 \hspace{3mm} Induced RNNG   & 95.2 & 50.6  \\
 \hspace{3mm} Induced URNNG  & 89.9 & 55.1   \\    
Neural PCFG & 252.6 & 52.6 \\ 
 \hspace{3mm} Induced RNNG   & 95.8 & 51.4 \\
 \hspace{3mm} Induced URNNG  & 86.0 & 58.7   \\    
Compound PCFG (\textbf{best}) & 196.3 & 60.1\\
 \hspace{3mm} Induced RNNG   & 89.8 & 58.1 \\
 \hspace{3mm} Induced URNNG  & \bf 83.7 & \bf 66.9 \\
\midrule
\textbf{Our results} (\textit{Averaged} over several trials) & PPL ($\downarrow$) & $F_1$ ($\uparrow$) \\
\midrule
Compound PCFG (re-experimented, \textbf{average}) & $-$ & 54.0 \\
 \hspace{3mm} Induced RNNG   & 91.5 &  54.7 \\
 \hspace{3mm} Induced URNNG  & 85.4 & 57.8 \\ 
CPE-PLM (All PLMs + \texttt{Greedy}) & $-$ & 55.3 \\
 \hspace{3mm} Induced RNNG   & 86.3 & 55.0 \\
 \hspace{3mm} Induced URNNG  & \bf 81.3 & 57.2 \\
CPE-PLM (All PLMs + \texttt{Beam}) & $-$ & 55.7 \\
 \hspace{3mm} Induced RNNG   & 87.3 & 57.0 \\
 \hspace{3mm} Induced URNNG  & 82.0 & \bf 60.7 \\
\bottomrule
\end{tabular}
\caption{Experiments on training (U)RNNG with the trees induced by unsupervised parsers and CPE-PLM.
The upper section presents the results reported by \citet{kim-etal-2019-unsupervised} while the bottom shows the outcomes from our experiments.
The best numbers for each column of the respective sections are in \textbf{bold}.
We show that the (U)RNNGs trained with the trees induced by CPE-PLM attain better language modeling and parsing abilities compared to the cases of unsupervised parsers.
}
\label{table:table4}
\end{table}

In Table \ref{table:table4}, we present results from \citet{kim-etal-2019-compound} and our experiments on PTB.
Note that our results and ones from the previous study are not directly comparable, because we report the performance of each model \textit{averaged} over 4 different runs while \citet{kim-etal-2019-compound} utilize the \textit{best} instance.
From the experimental results, we confirm that the (U)RNNG models trained with CPE-PLM have better language modeling capability than those trained with other unsupervised parsers.
In particular, we attain the perplexity of 81.3 when leveraging our greedy ensemble algorithm for CPE-PLM, outperforming the strong baseline (Compound PCFG: 85.4).
Moreover, we succeed in obtaining a more powerful parsing model by training (U)RNNG with the aid of CPE-PLM (All PLMs + \texttt{Beam}). 
Using this, we achieve 60.7 in $F1$, 5 points higher than that of the original (55.7).

\begin{table}[t!]
\scriptsize
\setlength{\tabcolsep}{0.3em}
\begin{center}
\begin{tabular}{l c c c c c}
\toprule
Models / Tasks (Metric: Acc.) & SST2 & MR & SUBJ & TREC \\ 
\midrule
Tree LSTM \\
+ Right-branching trees & 85.72 & 83.37 & 94.80 & 94.50 \\
+ \textbf{CPE-PLM (All PLMs + Beam ensemble)} & 86.10 & \bf 83.62 & 94.85 & 94.75 \\ 
+ Supervised parser \cite{klein2003accurate} & \bf 86.70 & \bf 83.62 & \bf 95.12 & \bf 95.05 \\ 
\bottomrule
\end{tabular}
\end{center}
\caption{Text classification with Tree LSTMs.
We observe that CPE-PLM-oriented parses outperform right-branching trees but are inferior to silver-standard trees.
All the results are averaged over four different runs.
}
\label{table:table5}
\end{table}

\subsection{Text Classification using Tree LSTM}

Recursive neural network (RvNN; \citet{socher-etal-2013-recursive,tai-etal-2015-improved}) is a type of neural architecture, whose composition order is determined by an input tree structure.
In spite of RvNNs' strong performance on several sentence-level tasks and robust linguistic motivation on which they were invented, the usage of RvNNs is generally restricted due to their reliance on gold/silver-standard trees.\footnote{We use the term \textit{silver-standard} trees to indicate parse trees predicted by sophisticated supervised parsers.}
We here attempt to mitigate this limitation by taking advantage of CPE-PLM.
To this end, we conduct experiments on four text classification tasks with Tree LSTMs: the target tasks are SST2 \cite{socher-etal-2013-recursive}, MR \cite{pang2005seeing}, SUBJ \cite{pang2004sentimental}, and TREC \cite{li2002learning}.
We use a variant of Tree LSTM \cite{kim2019dynamic} whose leaf nodes are processed by a separate LSTM in advance.
We inject three distinct types of trees---right-branching trees, which are a strong heuristic-based approach in English, the trees induced by CPE-PLM (with \texttt{Beam}), and those generated by a supervised parser---into the model and evaluate their impact on the final performance.

In Table \ref{table:table5}, we present the accuracy of diverse Tree LSTM instances on four tasks.
Although the absolute difference in accuracy between the instances is marginal, we discover a clear pattern that silver-standard trees (ones from supervised parsers) are always the most helpful while the parses induced by CPE-PLM rank second, outperforming right-branching trees.
This outcome supports our claim that CPE-PLM can be an attractive option when supervised parsers are not available.

On the other hand, we find that the performance of Tree LSTMs is not that sensitive to their tree inputs, which was similarly observed by \citet{shi-etal-2018-tree}.
However, we highlight that the trees closer to their gold-standard counterparts are more beneficial across all the tasks considered.
We leave as future work the application of CPE-PLM to advanced tree models that are more input structure-sensitive.

\begin{table}[t!]
\scriptsize
\centering
\begin{tabular}{l c c c c c}
\toprule
\multirow{2}{*}{CPE-PLM configurations} & \multicolumn{5}{c}{Used proportion of validation set} \\
& 1\% & 2\% & 5\% & 10\% & 100\% \\
\midrule
\bf All PLMs \\
+ \texttt{Greedy} & 49.4 & 49.9 & 52.7 & 54.3 & 55.3 \\
\hspace{3mm} Relative loss (-) & 5.9 & 5.3 & 2.5 & 0.9 & - \\
+ \texttt{Beam} & 51.3 & 49.8 & 51.8 & 52.9 & 55.7 \\
\hspace{3mm} Relative loss (-) & 4.5 & 6.0 & 4.0 & 2.9 & - \\
\bottomrule
\end{tabular}
\caption{Relative performance loss of CPE-PLM on PTB with regard to the proportion of the validation set used. 
We obtain reasonable performance only with 1\% (17 examples) of the validation set.}
\label{table:table6}
\end{table}

\begin{table}[t!]
\scriptsize
\centering
\setlength{\tabcolsep}{0.15em}
\begin{tabular}{l c c c c c c c c c}
\toprule
\multirow{2}{*}{Models} & \multicolumn{9}{c}{Number of used annotations} \\
& 1 & 2 & 5 & 10 & 17 (1\%) & 2\% & 5\% & 10\% & 100\% \\
\midrule 
\bf CPE-PLM (All PLMs) \\
+ \texttt{Greedy} & \textbf{46.2} & \textbf{48.4} & \textbf{49.9} & 49.1 & 49.4 & \textbf{49.9} & \textbf{52.7} & 54.3 & 55.3 \\
+ \texttt{Beam} & 45.4 & 45.9 & 47.7 & \textbf{49.6} & \textbf{51.3} & 49.8 & 51.8 & 52.9 & 55.7 \\
\midrule
\bf Supervised (Benepar) & - & 11.6 & 12.5 & 14.0 & 17.0 & 31.1 & 50.2 & \textbf{71.4 }& \textbf{92.2}\\
\bottomrule
\end{tabular}
\caption{Comparison between CPE-PLM and a supervised parser (Benepar) in few-shot settings.}
\label{table:table7}
\end{table}

\section{Discussion} \label{sec: discussion}

So far, we have focused on verifying the utility of CPE-PLM through the lens of (1) its improved parsing performance and (2) the effectiveness of its output trees for downstream tasks.
In this section, we conduct in-depth analysis on the limitations of the current form of CPE-PLM and propose countermeasures to alleviate the problems.

\paragraph{Reliance on the validation set.}

CPE-PLM is training-free, but it exploits gold-standard trees from the validation set to decide the best combination of attention heads ($g_{(m,n)}$).
Although we allow this configuration in this work to have a fair comparison with some previous work on unsupervised parsing \cite{kim-etal-2019-compound} that also made use of the validation set to optimize hyperparameters, 
it is always better to reduce such reliance as argued in the few-shot classification literature \cite{perez2021true}.
Therefore, we here attempt to examine the robustness of CPE-PLM with respect to the number of data instances from the validation set.
Specifically, we conduct a controlled experiment where CPE-PLM is provided with only a limited proportion of the validation set.
In Table \ref{table:table6}, we confirm that CPE-PLM only loses roughly five points in performance when just 17 (1\%) gold standard trees are available, implying that they work quite well even with a limited number of validation trees.

Furthermore, to showcase the data-efficiency of CPE-PLM in few-shot settings, we compare the performance of CPE-PLM and an off-the-shelf supervised parser (Benepar; \citet{kitaev-klein-2018-constituency}) in few-shot settings.\footnote{Specification on training a supervised parser in few-shot settings can be found in Appendix \ref{sec: details on training few-shot parsers}.}
From Table \ref{table:table7}, we discover that CPE-PLM shows much better performance than the normal parser in extreme cases where few dozen trees are provided.
When trees more than 10\% of the validation set are available, the supervised parser starts to outperform CPE-PLM. 

\begin{table}[t!]
\scriptsize
\centering
\setlength{\tabcolsep}{0.4em}
\begin{tabular}{l c c c}
\toprule
 \multirow{2}{*}{Models} & \multicolumn{2}{c}{$F1$ w/ ensemble ($\uparrow)$} & Inference \\
& \texttt{Greedy} & \texttt{Beam} & time ($\downarrow$) \\
\midrule
\textbf{Unsupervised parsers/CPE-PLM} \\
Compound PCFG \cite{kim-etal-2019-compound} & \multicolumn{2}{c}{55.2} & 31 min. \\
CPE-PLM (All PLMs) & 55.3 & 55.7 & 27 min. \\
\midrule
\textbf{Parsers trained with induced trees} \\
Distance \cite{shen-etal-2018-straight} & 53.8 & 55.0 & 36 sec. \\
Benepar \cite{kitaev-klein-2018-constituency} & \bf 56.6 & \bf 59.3 & \bf 32 sec. \\
\bottomrule
\end{tabular}
\caption{Training normal parsers with supervision from the trees induced by CPE-PLM. We show that it is viable to build a much faster parser while preserving (or even boosting) the performance of CPE-PLM by relying on existing techniques for supervised parsing.}
\label{table:table8}
\end{table}

\paragraph{Issues on the execution time.}

As identified in Table \ref{table:table8}, where we estimate the accuracy and execution time of different approaches on PTB, CPE-PLM and Compound PCFG are still too slow to be readily utilized, compared to supervised counterparts which are generally highly optimized.
To relieve this inefficiency, we propose to exploit normal parsers \cite{shen-etal-2018-straight,kitaev-klein-2018-constituency} by training them with the trees generated by CPE-PLM if a suitable amount of gold annotations are not available for supervision.
In Table \ref{table:table8}, we demonstrate that it is possible to transfer syntactic knowledge from PLMs to supervised parsers without loss of accuracy, while significantly reducing the execution time at the same time.
We even obtain performance gain in some cases, achieving nearly 60 in $F1$ score.
We expect that this direction can be particularly useful for low-resource languages for which it is hard to collect gold annotations.

\section{Conclusion}

In this paper, we introduce two ensemble methods and multi-PLM configurations for Constituency Parse Extraction from Pre-trained Language Models (CPE-PLM). 
We demonstrate that the performance of CPE-PLM can be competitive with that of unsupervised parsers with the aid of the proposed approaches, and that the parses induced by CPE-PLM are practically useful in several applications where parse trees are required as input.
We also propose solutions for mitigating some inherent limitations of CPE-PLM.
We anticipate that its potential will be further greater in the near future with the introduction of more sophisticated PLMs.

\section*{Acknowledgements}
We would like to thank anonymous reviewers for their fruitful feedback.
This work was supported by Institute of Information \& communications Technology Planning \& Evaluation (IITP) grant funded by the Korea government(MSIT) (No.2020-0-01373, Artificial Intelligence Graduate School Program (Hanyang University)).
This work was supported by the National Research Foundation of Korea(NRF) grant funded by the Korea government(MSIT) (No.2022R1F1A1074674). 

\bibliography{custom}

\clearpage

\appendix

\section {Appendix: Details on Training Few-shot Parsers} \label{sec: details on training few-shot parsers}

Following \citet{shi2020role}, we train a supervised parser (\textbf{Benepar}; \citet{kitaev-klein-2018-constituency}) in few-show learning settings to provide a robust baseline for our experiments.
To be specific, we leverage the official code and hyperparameters of the parser obtained from \url{https://github.com/nikitakit/self-attentive-parser}.
Given a designated number of parses from the PTB validation set, we utilize 90\% of them as the training set while the remaining 10\% are used as the real validation set.
We train the parser for 100 epochs, similar to \citet{shi2020role}.
Compared against the experimental results reported from \citet{shi2020role}, our few-shot parsers show relatively weaker performance.
We conjecture this gap comes from the methods \citet{shi2020role} exploited to boost their performance.
For instance, (1) they further pre-trained their word embeddings on sentences from PTB and (2) utilized data augmentation and self-training techniques, all of which are not applied in  our case.

\end{document}